\documentclass{article}

\usepackage{PRIMEarxiv}

\usepackage[utf8]{inputenc} % allow utf-8 input
\usepackage[T1]{fontenc}    % use 8-bit T1 fonts
\usepackage{hyperref}       % hyperlinks
\usepackage{url}            % simple URL typesetting
\usepackage{booktabs}       % professional-quality tables
\usepackage{amsfonts}       % blackboard math symbols
\usepackage{nicefrac}       % compact symbols for 1/2, etc.
\usepackage{microtype}      % microtypography
\usepackage{lipsum}
\usepackage{fancyhdr}       % header
\usepackage{graphicx}       % graphics
\usepackage{tabularx}
\usepackage{multirow}
\usepackage{pifont}
\usepackage{amsmath}

\usepackage{array}
\usepackage{pdflscape}
\usepackage{enumitem}
\graphicspath{{media/}}     % organize your images and other figures under media/ folder

\title{Text2Seg: Remote Sensing Image Semantic Segmentation via Text-Guided Visual Foundation Models
}

\author{
  Jielu Zhang$^*$ \\
  University of Georgia \\
  \texttt{jz20582@uga.edu} \\
   \And
  Zhongliang Zhou$^*$ \\
  University of Georgia \\
  \texttt{zz42551@uga.edu} \\
   \And
  Gengchen Mai$^{\dagger}$ \\
  University of Texas at Austin \\
  \texttt{gengchen.mai@austin.utexas.edu} \\
   \AND
  Mengxuan Hu \\
  University of Virginia \\
  \texttt{qtq7su@virginia.edu} \\
   \And
  Zihan Guan \\
  University of Virginia \\
  \texttt{bxv6gs@virginia.edu} \\
   \And
  Sheng Li \\
  University of Virginia \\
  \texttt{shengli@virginia.edu} \\
   \And
  Lan Mu$^{\dagger}$ \\
  University of Georgia \\
  \texttt{mulan@uga.edu} \\
}

\begin{document}
\maketitle

\begin{abstract}

Remote sensing imagery has attracted significant attention in recent years due to its instrumental role in global environmental monitoring, land usage monitoring, and more. As image databases grow each year, performing automatic segmentation with deep learning models has gradually become the standard approach for processing the data. Despite the improved performance of current models, certain limitations remain unresolved. Firstly, training deep learning models for segmentation requires per-pixel annotations. Given the large size of datasets, only a small portion is fully annotated and ready for training. Additionally, the high intra-dataset variance in remote sensing data limits the transfer learning ability of such models. Although recently proposed generic segmentation models like SAM have shown promising results in zero-shot instance-level segmentation, adapting them to semantic segmentation is a non-trivial task. To tackle these challenges, we propose a novel method named \textbf{Text2Seg} for remote sensing semantic segmentation. Text2Seg overcomes the dependency on extensive annotations by employing an automatic prompt generation process using different visual foundation models (VFMs), which are trained to understand semantic information in various ways. This approach not only reduces the need for fully annotated datasets but also enhances the model's ability to generalize across diverse datasets. Evaluations on four widely adopted remote sensing datasets demonstrate that Text2Seg significantly improves zero-shot prediction performance compared to the vanilla SAM model, with relative improvements ranging from 31\% to 225\%. Our code is available at \url{https://github.com/Douglas2Code/Text2Seg}.

\end{abstract}

% keywords can be removed

\section{Introduction}

\begin{figure*}[h]
  \centering
  \includegraphics[width=\textwidth]{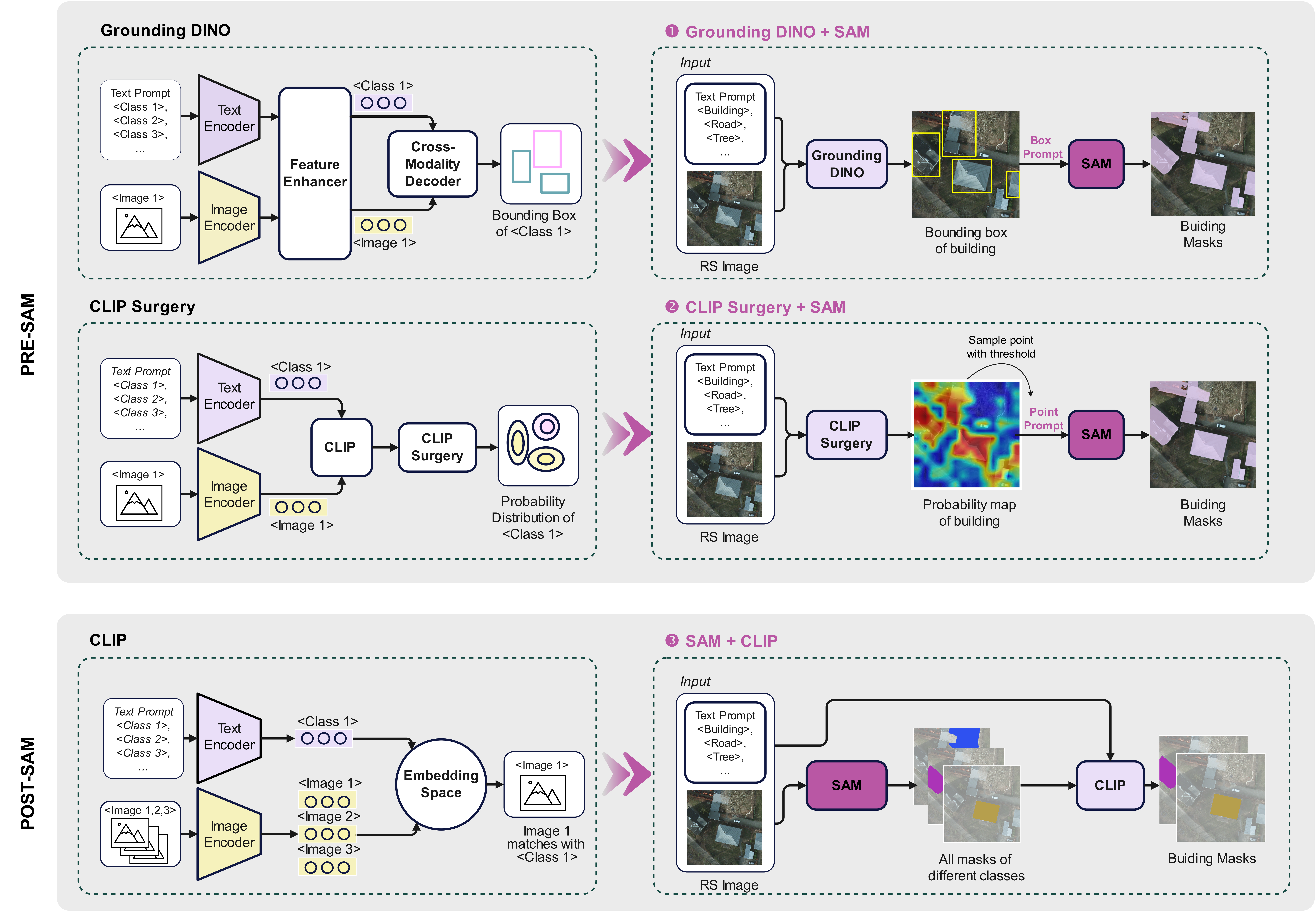}
  \caption{The overall structure of Text2Seg. This pipeline consists of three methods developed for guiding the SAM model for remote sensing semantic segmentation. \ding{182} Grounding DINO + SAM: a text prompt is used as input for Grounding DINO, which generates bounding boxes and input into SAM to produce a semantic segmentation map. \ding{183} CLIP Surgery + SAM: a text prompt is input into CLIP Surgery, yielding a probability map for that class. This probability map is sampled to create point prompts for SAM, which then generates semantic segmentation masks. 
  \ding{184} SAM + CLIP: We first utilize SAM to generate all available segmentation masks and then employ CLIP to compare their semantic similarity with the text prompt to identify masks of interest. 
  Apart from these three methods, we combine \ding{182} and \ding{183} to formulate the fourth method. We also combine all there methods (\ding{182}, \ding{183}, and \ding{184}) to formulate the fifth method.}
  \label{figure1}
\end{figure*}

Remote sensing tasks have become increasingly essential for applications such as environmental monitoring, land use estimation, and geospatial object detection \cite{yuan2021review}. These tasks depend heavily on the accurate segmentation of satellite images \cite{cong2022satmae,fuller2024croma}, UAV (Unmanned Aerial Vehicles) images \cite{lyu2020uavid}, and airborne sensor images \cite{sun2018fully}. The goal is to segment geospatial objects based on their semantic meaning, such as identifying grass, trees, buildings, water bodies, or vehicles. Traditionally, segmentation models have been developed separately for different datasets to handle discrepancies from varied sensors. However, these models often lack robustness and generalizability when applied to data from different geographic regions, times of the year, or types of remote sensing sensors \cite{he2022swin}.

In parallel with advancements in remote sensing, the development of large-scale foundational models for general tasks has seen significant progress. These models, particularly large language models, have shown remarkable zero-shot generalization capabilities across diverse applications, including public health, education, global warming estimation, disaster response, sustainability index prediction, autonomous spatial analysis, geographical localization, and medical assistance \cite{gptbubeck2023sparks, biswas2023potential, biswas2023role, zhou2024img2loc, surameery2023use, hu2023geo, manvi2024geollm, li2023autonomous}. Inspired by these achievements, researchers have begun creating general-purpose models in the visual learning domain. Initial efforts have focused on pre-training approaches that generate semantically rich numerical representations or linguistic descriptions of images \cite{li2022blip}. More recently, the research community has explored foundational models for semantic segmentation tasks, which require a detailed understanding of the geometric structure of images \cite{samkirillov2023segment}.

The Segment Anything Model (SAM) \cite{samkirillov2023segment}, introduced by Meta AI Research, is a pioneering foundational model for object segmentation. Trained on over 1 billion masks across 11 million images, SAM can perform zero-shot segmentation in diverse scenarios using visual prompts as guidance. This unique capability makes SAM particularly suitable for remote sensing imagery, where labeled datasets are often sparse and heterogeneous. However, using SAM directly for remote sensing image segmentation in a zero-shot setting presents significant challenges. The vanilla SAM segments each object into distinct masks, which is impractical for semantic segmentation tasks requiring masks with the same label for similar objects. Although fine-tuning SAM with adapter tuning shows potential for specific applications, it demands additional training data and resources, which may be scarce. Moreover, fine-tuned SAM models could be vulnerable to backdoor attacks, raising safety concerns \cite{guan2024badsam}.

To address these challenges, we propose a novel method named \textbf{Text2Seg} for remote sensing semantic segmentation. Recognizing that the vanilla SAM cannot perform semantic segmentation without an input prompt, we designed an automatic prompt generation process using different visual foundation models (VFMs) trained to understand semantic information in various ways. Text2Seg generates points or bounding boxes that narrow down the areas where SAM makes predictions or assist in filtering SAM's predictions based on specific text prompts. This approach reduces dependency on extensive per-pixel annotations and enhances the model's ability to generalize across diverse datasets. Evaluations on four widely adopted remote sensing datasets show that Text2Seg significantly improves zero-shot prediction performance compared to the vanilla SAM model. Since our method does not involve any additional training process, it is highly versatile and potentially applicable to various scenarios where labeled images are difficult to obtain.

\section{Related work}
\subsection{Remote Sensing Image Segmentation}
In recent years, remote sensing techniques have been widely used in Earth observation surveys, land use monitoring, and environmental protection \cite{gholizadeh2016comprehensive, angelopoulou2019remote}. The number of collected images from different sensors piles up every day, requiring experts to extract and analyze the information. To alleviate this human burden, there is a growing trend to leverage deep learning models for automatic remote sensing segmentation tasks. Many novel model architectures have been developed for remote sensing image segmentation, including CNN-based \cite{alam2021convolutional, kampffmeyer2016semantic}, Transformer-based \cite{he2022swin, wang2022unetformer}, and hybrid architectures \cite{zhang2022transformer, wang2019improved}. Despite improved performance, a persistent limitation of these models is that those trained on one dataset usually perform poorly on other datasets due to differences in sensors and dataset class distribution.

\subsection{Prompting for Foundation Models}
In the past two years, we have witnessed a surge in the development of foundation models. Led by OpenAI's GPT \cite{gptbubeck2023sparks} model family, foundation models have largely taken over the field of NLP with their superior performance in numerous downstream tasks in zero-shot or few-shot scenarios. As current foundation models demonstrate better-than-human performance on some tasks, adapting these models to real-world applications has become a hot research area. One particularly interesting technique is prompt engineering \cite{sahoo2024systematic}. Initially designed for large language models, prompt engineering leverages task-specific instructions to enhance model efficacy without further tuning the model parameters. With careful design of the prompts, it has been shown that the performance of foundation models can be easily boosted without computational overhead. The success of prompt engineering in NLP has inspired the exploration of developing similar strategies for visual foundation models \cite{gu2023systematic}.

\section{Methods}
\begin{figure}[bt]
  \centering
  \includegraphics[width=0.5\linewidth]{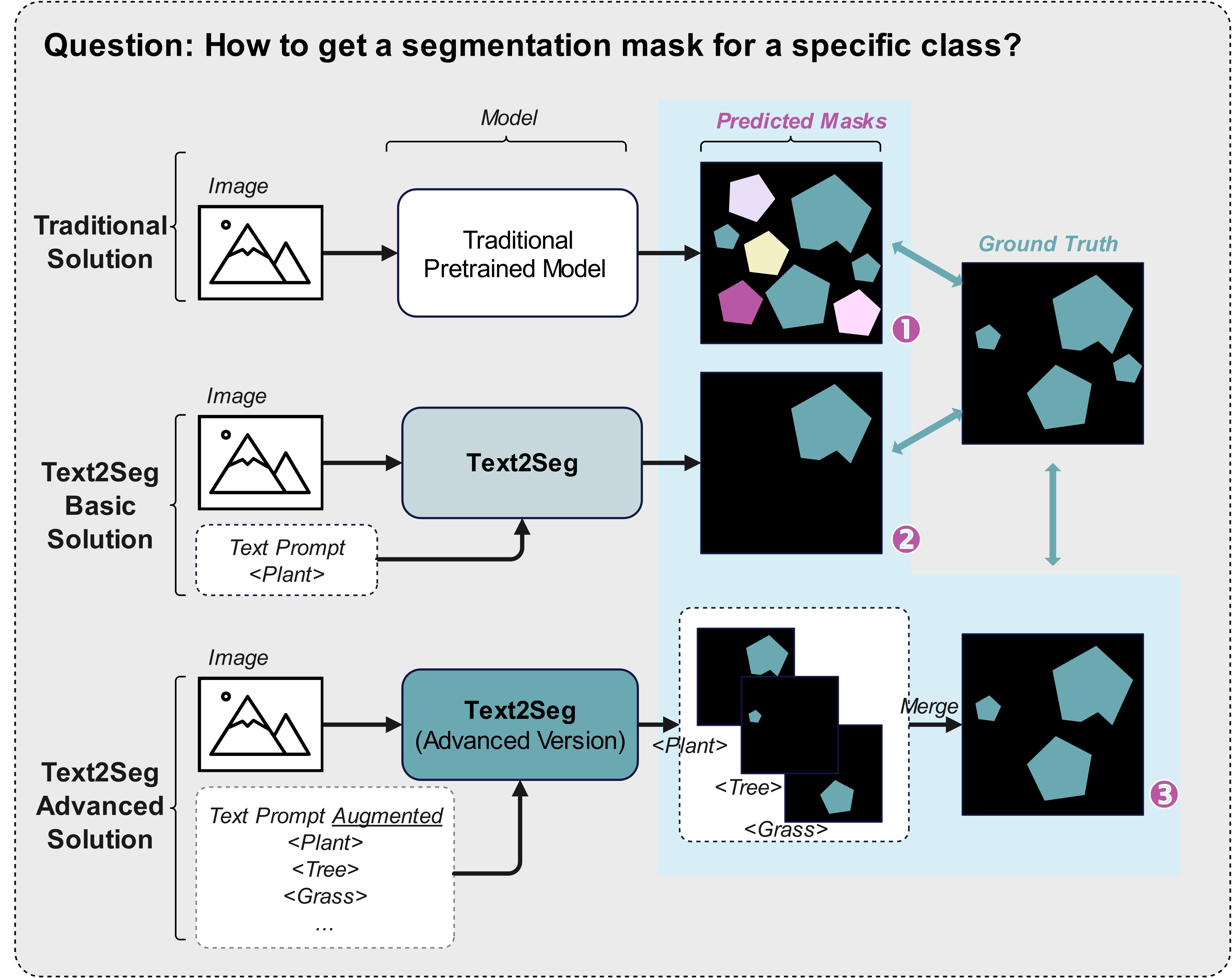}
  \caption{Comparison between traditional semantic segmentation models and our text-conditioned Text2Seg semantic segmentation model. Additionally, advanced Text2Seg allows for text prompt augmentation.}
  \label{figure2}
\end{figure}

\begin{figure*}[h]
  \centering
  \includegraphics[width=\textwidth]{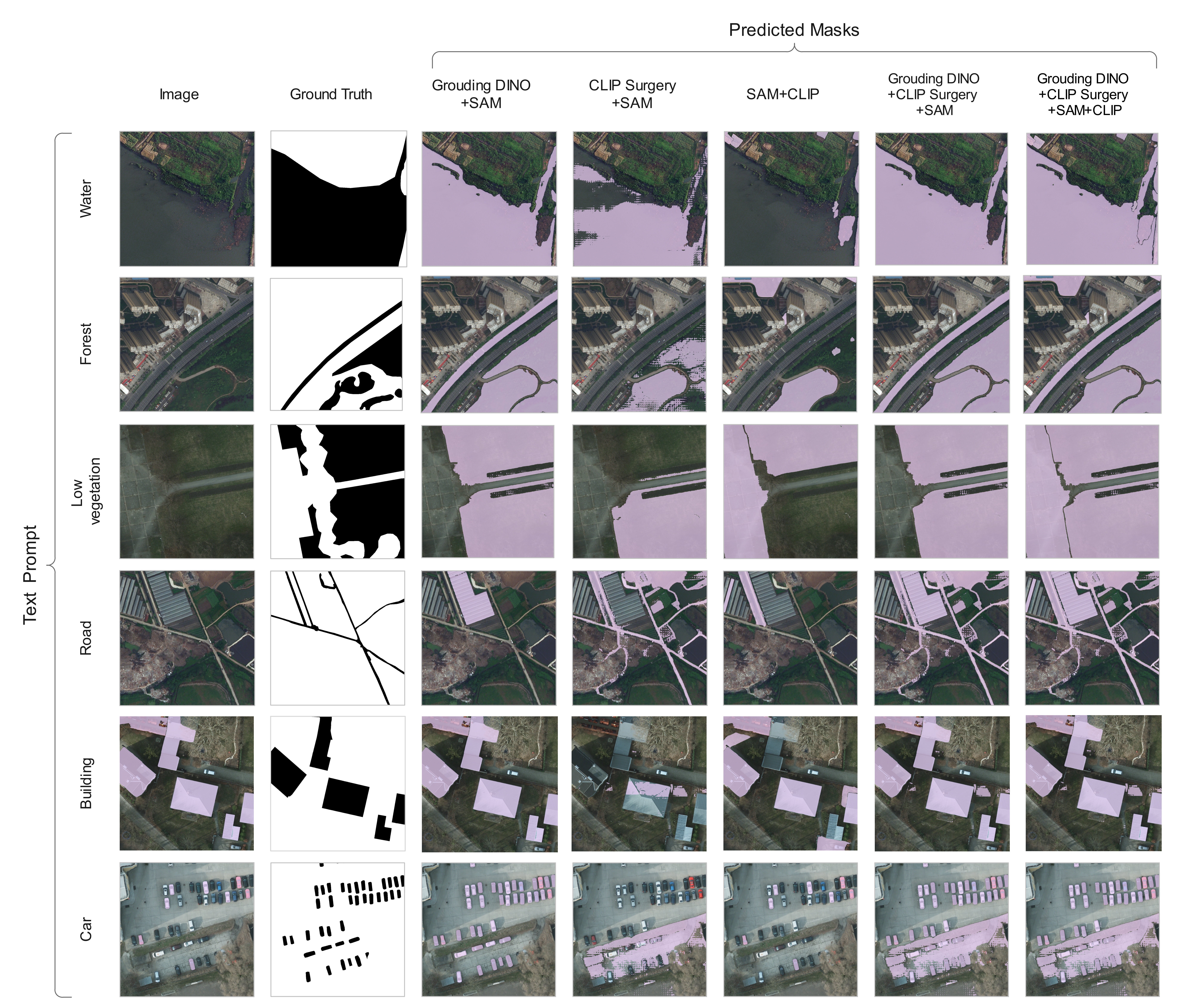}
  \caption{Predicted Masks from Text2Seg. This figure provides a qualitative demonstration of our proposed method across different classes in four datasets. The first column displays the original image, the second column shows the ground truth mask with the targeted class in black and other regions in white, and the subsequent columns present various combinations of results from our pipeline.}
  \label{figure3}
\end{figure*}
Given the limited availability of fully annotated remote sensing imagery for training and the constraints of transfer learning due to the high variance across different sensor data, we developed a highly general and efficient training-free methodology to address this issue by leveraging the strengths of several Visual Foundation Models(VFMs). To illustrate our approach, we first introduce the foundation models (FMs) integrated into our framework, including the Segment Anything Model (SAM)$(\delta)$\cite{samkirillov2023segment}, Grounding DINO$(\gamma)$ \cite{gdinoliu2023grounding}, CLIP$(\varphi)$ \cite{radford2021learning}, and CLIP Surgery$(\nu)$\cite{clipsli2023clip}. Then, we describe how our method is developed for incorporating them into a unified system designed for remote sensing semantic segmentation tasks \cite{wang2022unetformer}.

\begin{itemize}[leftmargin=*]
    \item \texttt{SAM$(\delta)$}: The Segment Anything model$(\delta)$ is designed to address the challenge of segmenting and identifying a wide variety of objects in images, regardless of their context or the dataset they originate from. Designed and trained with a promptable approach, the model can effortlessly transfer to new image distributions and tasks in a zero-shot fashion, utilizing points or boxes as prompts.
    
    \item \texttt{Grounding DINO$(\gamma)$}: The Grounding DINO model $(\gamma)$\cite{gdinoliu2023grounding} is an open-set object detection model. It can take human inputs like category names or referring expressions along with the images and output bounding boxes that detect arbitrary objects.
    
    \item \texttt{CLIP$(\varphi)$}: The Contrastive Language-Image Pre-training (CLIP)$(\varphi)$ model \cite{clipradford2021learning} is a multimodal foundation model pretrained with text-image pairs. The contrastive pretraining paradigm enables the model to align the embedding space between the text and image modalities, facilitating zero-shot image classification.
    
    \item \texttt{CLIP Surgery$(\nu)$}: The CLIP Surgery (CLIPS)$(\nu)$ \cite{clipsli2023clip} is an explanation method specifically designed for the CLIP model. For each input image and text pair, it examines the attention maps from the original CLIP's image encoder and finds the most relevant regions in the images, outputting a probability distribution of each pixel corresponding to the text concept.

\end{itemize}

\subsection{Text2Seg}
In traditional semantic segmentation settings, researchers usually design and train a model to simultaneously segment the entire image $I$ into various classes $\{C_1,C_2,\cdots,C_n\}$, as illustrated in the top row of Figure \ref{figure2}. However, for many applications, only a few classes $C_k$ are of interest and useful. Our Text2Seg model innovatively redesigns the semantic segmentation task as a conditioned task, allowing the user to choose which class they would like to segment, resulting in final prediction masks $y_k$ only for the designated class $C_k$ as illustrated in the second row of Figure \ref{figure2}. We define this problem as: Given an image $I$, a Text2Seg model$ (f:I \rightarrow y_k) $ is developed to segment objects attributed to a certain class $C_k$. We achieved this by integrating both the image $I$ and the desired class text concept $T_k$ as input for our model. Specifically, we examine three approaches that can be roughly categorized into two families: pre-SAM and post-SAM modules.

\subsubsection{Pre-SAM Module}
For the pre-SAM approach, Visual Foundation Models (VFMs) are applied prior to the SAM model to identify a visual prompt $V_k$ in either bounding box $(\square)$ or point format $(\cdot)$. These visual prompts $V_k$ are then input into SAM $\delta$ along with the original image to obtain the final prediction mask $y_k$. Within this approach, we explore generating either bounding boxes or points as prompts for SAM. For bounding box generation $(\square)$, we utilize the state-of-the-art open-set recognition Grounding DINO model $\gamma$, as illustrated in Figure \ref{figure1}. Grounding DINO $\gamma$ takes both images $I$ and a text prompt $T_k$, representing the class of interest, as input. The images $I$ and text $T_k$ are encoded using separate text $\mathcal{E}_{T}$ and image encoder $\mathcal{E}_{I}$. Following this, a feature enhancer employs deformable self-attention for image features $e_I$ and regular self-attention for text features $e_{T_k}$ to enhance the image and text features. The query selection mechanism guides the decoder $\mathcal{D}_{I}$ in identifying object locations in the image and assigning appropriate labels based on text descriptions. The cross-modality decoder $\mathcal{D}_{I}$ processes the fused features $e_{IT_{k}}$ and queries through a series of attention layers and feed-forward networks before producing the final bounding boxes $(\square)$. These bounding boxes $(\square)$ then serve as the input prompt $V_k$ for the SAM model $\delta$.

Another method in this module is to generate point prompts $(\cdot)$ to guide SAM $\delta$ for semantic segmentation using the CLIP Surgery method $\nu$. The CLIP Surgery method $\nu$ takes both images $I$ and a text prompt $T_k$ as input for the CLIP image encoder $\mathcal{E}_{I}$ and text encoder $\mathcal{E}_{T}$. It modifies the attention calculation step and replaces the feed-forward network with a designed dual-path approach. CLIP Surgery $\nu$ generates probability distribution maps $S$ for each pixel in the image $I$ with respect to a certain class $C_k$. This probability map $S$ refers to the similarity between pixels in the image $I$ and a certain text (class) $T_k$. A higher similarity score corresponds to a better alignment between the text prompt $T_k$ and the image $I$. We sample a predefined number of points from the maps at a threshold $t$ and use these points as input prompts $V_k$ for the SAM model $\delta$.
\raggedbottom
\subsubsection{Post-SAM Module}
In the post-SAM module, we aimed to refine the predictions of the original SAM $\delta$ by identifying objects most related to the input class $C_k$. For this purpose, we leverage the CLIP $\varphi$ model. Initially, grid points $(\cdot)$ are used as the input prompt for the SAM model $\delta$, generating a gallery of object instances $\{O_1,O_2,\cdots,O_n\}$. Subsequently, the CLIP image encoder $\mathcal{E}_{I}$ translates all these objects $\{O_1,O_2,\cdots,O_n\}$ into embeddings $\{e_{O_1},e_{O_2},\cdots,e_{O_n}\}$. These embeddings are then compared with the class embedding $e_{T_k}$ generated by the CLIP text encoder $\mathcal{E}_{T}$ to determine which objects $e_{O_n}$ are most similar to the target class. We set the combination of these selected objects ${e_{O_k}}$ to our final segmentation result for the current class $C_k$.

We also investigated the performance implications of integrating various methods within these two modules. In the first exploration, we combined the Grounding DINO-based pre-SAM method $\hat{\gamma}$ with the CLIP Surgery-based method $\hat{\nu}$ to evaluate the potential of prompt-guided SAM $\delta$. In the second exploration, we combined all three methods to analyze how the advantages and disadvantages of each interact. In total, five methods are included in our experiment.

\subsection{Text Prompt Augmentation}
Prompt engineering has proven effective in language foundation models. To explore its effectiveness for visual foundation models, understand the differences in applying prompt engineering between these two domains, and evaluate its applicability to remote sensing datasets, we designed a text prompt engineering pipeline. In the original version, we used the class label as the text prompt for all five methods to assist SAM in segmenting certain classes. In the prompt-engineered version, we augmented each class label with up to ten synonyms generated by GPT-4 and input these augmented text prompts into the developed model to enhance semantic segmentation. Taking class ``Plant'' as an example, the augmented prompts are ``Tree'',  ``Grass'', and `Plant''. For Grounding DINO+SAM, the bounding boxes will be generated from these augmentation texts using Grounding DINO. We then merge these bounding boxes together to guide SAM for segmentation. For CLIP Surgery + SAM, we merge the generated point prompts from CLIP Surgery together to guide SAM for segmentation. For SAM + CLIP, a set of segmented objects will be generated from SAM. We then compare all these segmented objects with all augmented texts using CLIP to get the final segmentation result. This augmentation result will be demonstrated in Section \ref{sec:ablation} and detail for augmentation is illustrated in Figure \ref{figure2}.
\raggedbottom
\section{Experiments}
\subsection{Experiment Setup}

In our experiment, we compare our proposed method with the original SAM model on remote sensing segmentation tasks. Since the original SAM does not allow semantic segmentation, we generate a random point for each image and use this point as guidance to generate a segmentation mask. As SAM does not assume the class label for this generated mask, we compare each class's ground truth with the prediction and take the maximum value as the predicted score. We want to emphasize that this comparison is not entirely fair for our model but it provides a rough estimation of SAM's performance.

For our Text2Seg pipeline, we explore the best practices for combining multiple visual foundation models to generate visual prompts that guide the SAM model in performing semantic segmentation. We test our pipeline with different combinations, specifically: \ding{182} Grounding DINO + SAM, \ding{183} CLIP Surgery + SAM, \ding{184} SAM + CLIP, \ding{185} Grounding DINO + CLIP Surgery + SAM, and \ding{186} Grounding DINO + CLIP Surgery + SAM + CLIP. 

We evaluate these combinations on four different remote sensing datasets: UAVid\cite{lyu2020uavid}, Vaihingen\cite{rottensteiner2012isprs}, Potsdam\cite{rottensteiner2012isprs}, and LoveDA\cite{wang2021loveda}.

\begin{table*}
\centering
\caption{Zero-shot Semantic Segmentation Result on Remote Sensing Dataset. The bold number indicates the highest performance for a particular class across all methods (a row).}
\label{basic}
\resizebox{\textwidth}{!}{
\begin{tabular}{c|c|cc|cc|cc|cc|cc|cc}
\hline
\multirow{2}{*}{\textbf{Dataset}} & \multirow{2}{*}{\textbf{Class}} & \multicolumn{2}{c|}{\textbf{\begin{tabular}[c]{@{}c@{}}SAM\\ (Baseline)\end{tabular}}} & \multicolumn{2}{c|}{\textbf{GDINO+SAM}} & \multicolumn{2}{c|}{\textbf{CLIPS+SAM}} & \multicolumn{2}{c|}{\textbf{SAM+CLIP}} & \multicolumn{2}{c|}{\textbf{\begin{tabular}[c]{@{}c@{}}GDINO+CLIPS+\\ SAM\end{tabular}}} & \multicolumn{2}{c}{\textbf{\begin{tabular}[c]{@{}c@{}}GDINO+CLIPS+\\ SAM+CLIP\end{tabular}}} \\ \cline{3-14} 
 &  & \textbf{IoU} & \textbf{OA} & \textbf{IoU} & \textbf{OA} & \textbf{IoU} & \textbf{OA} & \textbf{IoU} & \textbf{OA} & \textbf{IoU} & \textbf{OA} & \textbf{IoU} & \textbf{OA} \\ \hline
\multirow{5}{*}{\textbf{UAVid}} & \textit{Building} & 0.147 & 0.495 & \textbf{0.655} & \textbf{0.859} & 0.147  & 0.658 & 0.349 & 0.737 & 0.620 & 0.857 & 0.468 & 0.833 \\
 & \textit{Road} & 0.468 & 0.803 & \textbf{0.590} & 0.877 & 0.056 & 0.797 & 0.204 & 0.802 & 0.577 & \textbf{0.883} & 0.299 & 0.812 \\
 & \textit{Tree} & 0.169 & 0.651 & 0.473 & \textbf{0.859}  & 0.033 & 0.716  & 0.292 & 0.798  & \textbf{0.474}  & \textbf{0.859}  & 0.376  & 0.852  \\
 & \textit{Low Vegetation} & 0.295 & 0.773 & \textbf{0.300}  & 0.784  & 0.067 & 0.758 & 0.211 & \textbf{0.822}  & 0.285 & 0.780  & 0.243 & 0.787  \\
 & \textit{Car} & 0.054 & 0.963 & \textbf{0.517} & \textbf{0.967}  & 0.014  & 0.832 & 0.148 & 0.929 & \textbf{0.517}  & \textbf{0.967}  & 0.216 & 0.927 \\ \cline{2-14} 
 & \textbf{Overall} & 0.226 & 0.802 & \textbf{0.507} & \textbf{0.869} & 0.063 & 0.752 & 0.241 & 0.817 \ & 0.494 & \textbf{0.869} & 0.320 & 0.842 \\ \hline
\multirow{6}{*}{\textbf{LoveDA}} & \textit{Building} & 0.047 & 0.776 & \textbf{0.357}& 0.649 & 0.071& 0.777 & 0.156 & \textbf{0.837} & 0.352 & 0.657 & 0.224 & 0.608 \\
 & \textit{Road} & 0.353 & \textbf{0.902} & \textbf{0.355} & 0.641 & 0.043 & 0.779 & 0.232 & 0.698 & 0.346 & 0.654 & 0.154 & 0.608 \\
 & \textit{Water} & \textbf{0.384} & \textbf{0.813} & 0.276 & 0.564 & 0.075 & 0.758 & 0.012 & 0.806 & 0.267 & 0.570 & 0.175 & 0.543 \\
 & \textit{Forest} & \textbf{0.227} & \textbf{0.814} & 0.157 & 0.399 & 0.050 & 0.779 & 0.020 & 0.777 & 0.162 & 0.418 & 0.136 & 0.414 \\
 & \textit{Agriculture} & 0.174 & 0.537 &\textbf{0.409} &0.517 &0.094 & \textbf{0.678}& 0.091 & 0.575 & 0.400 & 0.519 & 0.351 & 0.517 \\ \cline{2-14} 
 & \textbf{Overall} & 0.237 & 0.738 & \textbf{0.311} & 0.554 & 0.067 & \textbf{0.754} & 0.102 & 0.738 & 0.305 & 0.564 & 0.208 & 0.538 \\ \hline
\multirow{6}{*}{\textbf{Vaihingen}} & \textit{Impervious Surface} & 0.050 & 0.681 & \textbf{0.205} & 0.600 & 0.044 & 0.661 & 0.039 & \textbf{0.682} & 0.223 & 0.606 & 0.181 & 0.596 \\
 & \textit{Building} & 0.098 & 0.691 & \textbf{0.566} & 0.822 & 0.080 & 0.700 & 0.334 & 0.816 & 0.564 & \textbf{0.824} & 0.462 & 0.811 \\
 & \textit{Low Vegetation} & 0.104 & 0.751 & \textbf{0.221} & 0.687 & 0.098 & \textbf{0.756} & 0.053 & 0.751 & 0.210 & 0.698 & 0.198 & 0.669 \\
 & \textit{Tree} & 0.132 & 0.725 & 0.337 & 0.820 & 0.093 & 0.734 & 0.127 & 0.790 & \textbf{0.350} & \textbf{0.825} & 0.284 & 0.812 \\
 & \textit{Car} & 0.112 & 0.989 & \textbf{0.267} & \textbf{0.992} & 0.013 & 0.890 & 0.020 & 0.927 & 0.262 & \textbf{0.992} & 0.056 & 0.945 \\ \cline{2-14} 
 & \textbf{Overall} & 0.099 & \textbf{0.803} & 0.320 & 0.783 & 0.065 & 0.748 & 0.115 & 0.793 & \textbf{0.322} & 0.789 & 0.236 & 0.767 \\ \hline
\multirow{6}{*}{\textbf{Potsdam}} & \textit{Impervious Surface} & 0.254 & \textbf{0.717} & \textbf{0.352} & 0.665 & 0.133 & 0.636 & 0.084 & 0.646 & 0.328 & 0.680 & 0.245 & 0.645 \\
 & \textit{Building} & 0.197 & 0.721 & \textbf{0.663} & 0.812 & 0.131 & 0.698 & 0.433 & 0.798 & 0.643 & \textbf{0.820} & 0.410 & 0.800 \\
 & \textit{Low Vegetation} & 0.343 & 0.796 & \textbf{0.367} & 0.804 & 0.093 & 0.685 & 0.238 & 0.788 & 0.335 & \textbf{0.808} & 0.289 & 0.804 \\
 & \textit{Tree} & 0.200 & 0.781 & \textbf{0.339} & \textbf{0.879} & 0.050 & 0.735 & 0.148 & 0.818 & 0.341 & 0.876 & 0.271 & 0.850 \\
 & \textit{Car} & 0.081 & 0.965 & \textbf{0.619} & \textbf{0.991} & 0.017 & 0.806 & 0.086 & 0.918 & 0.617 & 0.990 & 0.280 & 0.937 \\ \cline{2-14} 
 & \textbf{Overall} & 0.215 & 0.802 & \textbf{0.468} & 0.830 & 0.085 & 0.712 & 0.197 & 0.794 & 0.453 & \textbf{0.835} & 0.299 & 0.807 \\ \hline
\end{tabular}
}
\end{table*}

\subsection{Dataset}
\subsubsection{UAVid dataset}
The UAVid dataset is a high-resolution UAV semantic segmentation dataset focusing on urban street scenes, providing two distinct spatial resolutions ($ 3840 \times 2160 $ and $ 4096 \times 2160 $) and encompassing eight classes, as detailed in \cite{lyu2020uavid}. Segmentation of UAVid images poses challenges due to their high spatial resolution, diverse spatial features, ambiguous categories, and complex scenes. We employed the officially provided set of 150 images to demonstrate our findings. In our experimental methodology, each image was padded and partitioned into eight $ 1024 \times 1024 $ pixel patches, adhering to the procedure described in \cite{wang2022unetformer}.
\subsubsection{LoveDA dataset}
The LoveDA dataset\cite{wang2021loveda} comprises 5987 fine-resolution optical remote sensing images (GSD 0.3 m), each measuring 1024 by 1024 pixels, and features 7 land cover categories: building, road, water, barren, forest, agriculture, and background\cite{wang2021loveda}. The dataset covers two scenes (urban and rural) collected from three Chinese cities: Nanjing, Changzhou, and Wuhan. This variety introduces considerable challenges due to the presence of multiscale objects, complex backgrounds, and inconsistent class distributions. 
\subsubsection{Vaihingen dataset}
The Vaihingen dataset\cite{rottensteiner2012isprs} consists of 33 high-resolution TOP image tiles, averaging $ 2494 \times 2064 $ pixels in size. Each tile features three multispectral bands (near-infrared, red, and green) and includes both a digital surface model (DSM) and a normalized digital surface model (NDSM) with a 9 cm ground sampling distance (GSD). The dataset contains five foreground classes (impervious surface, building, low vegetation, tree, and car) and one background class. In our experiments, we only used the TOP image tiles without the DSM and NDSM data. These image tiles were then cropped into patches measuring 1024 by 1024 pixels.
\subsubsection{Potsdam dataset}
The Potsdam dataset \cite{rottensteiner2012isprs} comprises 38 high-resolution TOP image tiles (with 5 cm GSD), each measuring $ 6000 \times 6000 $ pixels, and encompasses the same category information as the Vaihingen dataset. This dataset provides four multispectral bands (red, green, blue, and near-infrared) in addition to the DSM and NDSM. In our experiments, we utilized only the red, green, and blue bands, and cropped the original image tiles into patches of 1024 by 1024 pixels.

\subsection{Evaluation Metrics}

The evaluation metrics used in our experiments follow previous works \cite{wang2022unetformer}, which include overall accuracy (OA) and mean intersection over union (mIoU).

1. \textbf{Overall Accuracy (OA)}: This metric measures the proportion of correctly classified pixels over the total number of pixels $n$.
\[
\text{OA} = \frac{\sum_{i=1}^{n} TP_i}{\sum_{i=1}^{n} (TP_i + FP_i + FN_i + TN_i)}
\]
where \( TP \) are true positives, \( FP \) are false positives, \( FN \) are false negatives, and \( TN \) are true negatives.

2. \textbf{Mean Intersection over Union (mIoU)}: This metric measures the overlap between the predicted segmentation and the ground truth, averaged over all classes.
\[
\text{IoU} = \frac{TP}{TP + FP + FN} \quad \quad \text{mIoU} = \frac{1}{n} \sum_{i=1}^{n} \text{IoU}_i
\]

For the original SAM model, since it cannot generate segmentation results by class, we use the best score across classes to ensure a fair evaluation. Specifically, for each metric (OA, mIoU), we take the highest score obtained among all classes:
\[
\text{Best Score} = \max_{c \in \text{Classes}} \text{Metric}_c
\]
This approach provides SAM with the most favorable evaluation results given its class-agnostic segmentation capabilities.

\subsection{Result}

Based on our evaluation results (Table \ref{basic}), we observe consistent improvements over the original SAM for zero-shot remote sensing segmentation tasks. Specifically, on the UAVid dataset, our best pipeline improved the original SAM model's mIoU from 0.226 to 0.507, representing a 124\% relative improvement. On the LoveDA dataset, the mIoU increased from 0.237 to 0.311, a 31\% relative improvement. The Vaihingen dataset saw an increase from 0.099 to 0.322, a 225\% relative improvement. On the Potsdam dataset, the mIoU improved from 0.215 to 0.468, a 117\% relative improvement. These consistent improvements clearly demonstrate the advantage and robustness of our pipeline across different datasets.

However, we also observe that the performance improvements are not uniform across different datasets and classes. For example, the Vaihingen dataset experienced the highest relative improvement, while the LoveDA dataset showed the smallest. This suggests that disparities between datasets, caused by factors such as sensor types and resolution, may limit optimal performance. For different classes, our pipeline usually made significant improvements on classes like buildings or cars but smaller improvements on classes like low vegetation or water regions. We speculate that this is due to the SAM model's tendency to segment whole objects into smaller parts, which lowers accuracy for classes like vegetation or water regions. Additionally, the color of vegetation and water can vary depending on the sensor bands used, further contributing to lower performance for these classes.

Finally, we observe different performance improvements for the various pipelines we developed. The Grounding DINO + SAM pipeline showed the best improvement across all datasets, while the CLIP Surgery + SAM pipeline had the smallest improvement on a few datasets. We argue that the inferior results of CLIP Surgery are due to its ability to provide only a rough estimation of the class distribution. Consequently, sampling from this distribution may not yield an effective point prompt for SAM.

\section{Ablation Study} \label{sec:ablation}

\begin{table}[t]

\centering
\caption{Ablation Study Results Comparing Situations with and without Augmentation}

\vspace{-3mm}

\label{ablation}
\resizebox{0.5\textwidth}{!}{
\begin{tabular}{c|c|c|c}
\hline
\multirow{2}{*}{\textbf{Dataset}} & \multirow{2}{*}{\textbf{\begin{tabular}[c]{@{}c@{}}Text Prompt \\ Augmentation\end{tabular}}} & \multicolumn{2}{c}{\textbf{mIoU}} \\ \cline{3-4} 
 &  & \textbf{\begin{tabular}[c]{@{}c@{}}Pre-SAM\\ GDINO+SAM\end{tabular}} & \textbf{\begin{tabular}[c]{@{}c@{}}Post-SAM\\ SAM+CLIP\end{tabular}} \\ \hline
\multirow{2}{*}{\textbf{UAVid}} & \ding{55} Augmentation & 0.507 & 0.241 \\
 & \ding{51} Augmentation & 0.543 $\uparrow$ & 0.311 $\uparrow$ \\ \hline
\multirow{2}{*}{\textbf{LoveDA}} & \ding{55} Augmentation & 0.311 & 0.102 \\
 & \ding{51} Augmentation & 0.378 $\uparrow$ & 0.243 $\uparrow$ \\ \hline
\multirow{2}{*}{\textbf{Vaihingen}} & \ding{55} Augmentation & 0.320 & 0.115 \\
 & \ding{51} Augmentation & 0.327 $\uparrow$ & 0.217 $\uparrow$ \\ \hline
\multirow{2}{*}{\textbf{Potsdam}} & \ding{55} Augmentation & 0.468 & 0.197 \\
 & \ding{51} Augmentation & 0.492 $\uparrow$ & 0.244 $\uparrow$ \\ \hline
\end{tabular}}

\vspace{-5mm}
\end{table}
The uniqueness of our method lies in providing a surrogate for the original points or bounding box prompt engineering of SAM in a text format. This makes the selection of text pivotal for the final performance. To evaluate the effect of our text prompt augmentation pipeline, we conducted an ablation study by selecting two methods from our proposed toolbox and comparing the mIoU scores before and after the augmentation. Based on the results in Table \ref{ablation}, we can see that text augmentation improves the overall performance of both the Pre-SAM and Post-SAM models across different datasets, suggesting that the augmented text concepts from GPT-4 can help improve the understanding of the visual language models of the corresponding class, thus leading to better model performance.

\section{Conclusion}

In this paper, we propose that leveraging multiple visual foundation models enables their efficient repurposing for specific downstream tasks, such as zero-shot remote sensing semantic segmentation. Our findings suggest that while the SAM model effectively segments instances within the provided frame, generating segmentation masks for a specific category remains challenging. In contrast, other visual foundation models, like Grounding DINO and CLIP, exhibit a superior capacity to understand the semantic features of images, generating coarse-grained visual prompts. Integrating these models into a unified pipeline allows us to exploit their combined strengths. Our results on four different datasets demonstrate the effectiveness of our pipeline. Admittedly, compared to fully supervised models, there is still room for improvement. We reason that the inferior performance compared to supervised counterparts could be due to the lack of remote sensing images in the pretraining data distribution of these visual foundation models. Therefore, we advocate for the development of visual foundation models tailored to specific domains. We hope that our systematic evaluation and proposed pipeline will encourage additional research on the application of visual foundation models for diverse downstream tasks and stimulate the development of increasingly powerful models.

%Bibliography
\bibliographystyle{unsrt}  
\bibliography{references}

\end{document}